\documentclass[letterpaper, 10 pt, conference]{ieeeconf} 

\IEEEoverridecommandlockouts

\usepackage{url}
\usepackage{color}
\usepackage{graphicx}
\usepackage{subfig}
\usepackage{amsmath}
\usepackage[table,xcdraw]{xcolor}
\usepackage{adjustbox}

\usepackage{soul}

\title{\LARGE \bf
	Speeding-up Object Detection Training for Robotics with FALKON
}

\author{Elisa Maiettini$^{1,2,3}$, Giulia Pasquale$^{1,2}$, Lorenzo Rosasco$^{2,3}$ and Lorenzo Natale$^{1}$
	\thanks{$^{1}$ iCub Facility, Istituto Italiano di Tecnologia, Genoa, Italy}%
	\thanks{$^{2}$ Laboratory for Computational and Statistical Learning, Istituto Italiano di Tecnologia and Massachusetts Institute of Technology, Cambridge, MA}%
	\thanks{$^{3}$ Dipartimento di Informatica, Bioingegneria, Robotica e Ingegneria dei Sistemi, University of Genoa, Genoa, Italy}%
}

\begin{document}

	\maketitle
	\thispagestyle{empty}
	\pagestyle{empty}

	%%%%%%%%%%%%%%%%%%%%%%%%%%%%%%%%%%%%%%%%%%%%%%%%%%%%%%%%%%%%%%%%%%%%%%%%%%%%%%%%
	\begin{abstract}
		
		Latest deep learning methods for object detection provide remarkable performance, but have limits when used in robotic applications. One of the most relevant issues is the long training time, which is due to the large size and imbalance of the associated training sets, characterized by few positive and a large number of negative examples (i.e. background). Proposed approaches are based on end-to-end learning by back-propagation~\cite{ren2015_faster} or kernel methods trained with Hard Negatives Mining on top of deep features~\cite{girshick2014_rcnn}. These solutions are effective, but prohibitively slow for on-line applications.
		
		In this paper we propose a novel pipeline for object detection that overcomes this problem and provides comparable performance, with a 60x training speedup.
		Our pipeline combines (i) the Region Proposal Network and the deep feature extractor from~\cite{ren2015_faster} to efficiently select candidate RoIs and encode them into powerful representations, with (ii) the FALKON~\cite{falkon2018} algorithm, a novel kernel-based method that allows fast training on large scale problems (millions of points). We address the size and imbalance of training data by exploiting the stochastic subsampling intrinsic into the method and a novel, fast, bootstrapping approach.
		
		We assess the effectiveness of the approach on a standard Computer Vision dataset (PASCAL VOC 2007~\cite{pascal2010}) and demonstrate its applicability to a real robotic scenario with the iCubWorld Transformations~\cite{pasquale2016} dataset.
		
	\end{abstract}
	
	%%%%%%%%%%%%%%%%%%%%%%%%%%%%%%%%%%%%%%%%%%%%%%%%%%%%%%%%%%%%%%%%%%%%%%%%%%%%%%%%
	
	\section{INTRODUCTION}
	\label{sec:intro}
	
	Visual recognition and localization of objects is crucial for robotic platforms. This problem is known in Computer Vision as object detection, where the objects represented in an image are localized with a bounding box and then associated to a label.
	Deep learning has remarkably boosted the performance of the state of the art~\cite{everingham2015, russakovsky2015, coco, sermanet2014, girshick2014_rcnn, ren2015_faster, liu2015_ssd, dai2016} but training such systems require large amount of data and it is slow. For robotic applications it is desirable to use methods that allow robots to quickly adapt to the environment. 
	
	In previous work~\cite{pasquale2016, pasquale2016_frontiers} we proposed an interactive method that allows the robot to automatically acquire annotated images by interacting with a human and demonstrated that data acquired in this way allows training an object detection algorithm (i.e. Faster R-CNN~\cite{ren2015_faster}) with good accuracy~\cite{maiettini2017}.
	
	However, in \cite{maiettini2017} training was performed off-line as in~\cite{ren2015_faster}, i.e., by fine-tuning the model on the task at hand. Most latest architectures for object detection are learned end-to-end by gradient descent with back-propagation~\cite{ren2015_faster, liu2015_ssd, dai2016, redmon2016yolo9000}. This is achieved by formulating a single optimization problem, to learn, jointly, the three stages of the classical detection pipeline: extraction of candidate regions, feature encoding and regions classification (see~\cite{huang2016} for a comprehensive overview). This training protocol require massive computational resources and hours/days of training. In this paper we propose an object detection pipeline that can be trained on-line, (i.e. in seconds), without sacrificing detection performance. 
	
	We adopt an approach that is similar to Region-CNN~\cite{girshick2014_rcnn}, in which each stage of the pipeline is trained independently. This approach is more suitable for an on-line learning because the region proposals and the deep representation can be kept fixed while the final region classification is trained. For instance, Region-CNN uses Selective Search~\cite{uijlings2013} and a pre-trained CNN to get candidate regions and encode them into features, on top of which Support Vector Machines (SVMs) are trained.
	
	Training neural models for object detection is computationally demanding. This is due to the large number of candidate regions that are extracted from the training images, especially from the background. Typically a time consuming process is applied to extract a small number of negative examples on which to train the system. Such process aims at selecting a subset of negative examples that are meaningful (i.e. `hard') and whose size is comparable to the available positive examples, thus re-balancing the dataset. As an example training in Region-CNN uses the Hard Negative Mining procedure of~\cite{Sung1996} and it requires hours to complete.
	
	We compose our pipeline by first adopting the Region Proposal Newtork (RPN) presented in Faster R-CNN~\cite{ren2015_faster} for learning the candidate regions: this is a double-layer Convolutional Neural Network (CNN)
	that can be trained off-line in relatively short time and, since it is object-agnostic, can be used for multiple object detection tasks. Using another deep CNN we encode each region into a set of feature vectors. These feature vectors are then fed to a kernel-based classifier which is trained on-line. In doing so, for the first off-line part, we basically adopt the architecture of Faster R-CNN~\cite{ren2015_faster}, except that the RPN and the feature extractor CNN are learned on one task, and then re-used on different tasks.
	
	This choice is motivated by the literature suggesting that deep CNNs provide powerful and general features which can be used for multiple tasks~\cite{razavian2014, donahue2014, jia2014, girshick2014_rcnn, pasquale2016}. 
	
	For region classification we adopt FALKON~\cite{falkon2018}, a recent kernel-based method for large-scale datasets. We leverage on the stochastic sampling of the kernel centers performed by FALKON and propose an approximated version of the Hard Negative Mining procedure, to efficiently re-balance the training set. We show that depending on the desired computation time, this procedure can be tuned to subsample more or less extensively the training set without compromising performance. 
	
	We validate the pipeline on a subset of the PASCAL VOC 2007~\cite{pascal2010}, by comparing performance and training time with Faster R-CNN. Finally, we experiment with a real-world robotic application to demonstrate that the proposed approach allows learning a novel task ($10$ classes $\times$ $10$k images) in few seconds. To this end, we use the iCubWorld Transformations Dataset~\cite{pasquale2016}, a robotic benchmark for object recognition and detection.

	%%%%%%%%%%%%%%%%%%%%%%%%%%%%%%%%%%%%%%%%%%%%%%%%%%%%%%%%%%%%%%%%%%%%%%%%%%%%%%%%
	\section{RELATED WORK}
	
	\label{sec:rel_works}

	The problem of object detection can be naturally decomposed into two subtasks: (1) objects localization and (2) image classification. Historically many approaches have been proposed which address these tasks in different ways. They can be grouped as follows:\\
	
	\noindent{{\bf Grid-based object detectors.}}
	In this case a classifier is applied on a dense image grid, obtained using a Sliding Window paradigm~\cite{LeCun1989,Felzenszwalb2010} or a fixed stride. The work of LeCun et al.~\cite{LeCun1989} is one of the first where convolutional neural networks were applied in a Sliding Window fashion. More recently, other grid-based approaches have been proposed like SSD (Single-Shot MultiBox Detector)~\cite{liu2015_ssd} and YOLO (You Only Look Once)~\cite{redmon2016, redmon2016yolo9000}.\\
	
	\noindent{{\bf Region-based object detectors.}}
	Algorithms belonging to this group perform detection only on a set of ``candidate'' Regions of Interest (RoIs), selected with a separate process (see e.g., Region-CNN (R-CNN)~\cite{girshick2014_rcnn} and its optimizations Fast R-CNN~\cite{girshick15_fastrcnn}, Faster R-CNN~\cite{ren2015_faster}, Region-FCN~\cite{dai2016} and Mask R-CNN~\cite{He2017}). These architectures employ different methods for predicting RoIs. For example Region-CNN usually relies on Selective Search~\cite{uijlings2013}, while Faster R-CNN uses a dedicated CNN, called Region Proposal Network (RPN)~\cite{ren2015_faster}. The RPN is a double-layer Convolutional Neural Network, which is trained on the data to discriminate background from potential objects. The main advantage of this approach is that it provides a limited number of predicted RoIs with low computation time~\cite{ren2015_faster}.  
	
	Among these architectures we can distinguish between those which are trained end-to-end~\cite{dai2016,He2017,ren2015_faster} and those which need a multi-stage learning process~\cite{girshick2014_rcnn}.
	
	In this work we propose a pipeline that exploits the benefits of an RPN for RoIs prediction while relying on a two-stage training procedure to allow fast model adaptation to new tasks.\\
	
	\noindent{{\bf The problem of background selection.}}
	The problem with grid-based and region-based detectors is that they produce a large number of RoIs, most of which originate from the background. Because negative examples are computed from these RoIs, this leads to a training set that is large and highly unbalanced. The size of the training set makes learning computationally unfeasible, and the fact that positive examples are underrepresented inevitably bias the result of the training.
	
	In the literature, methods have been proposed to deal with these issues. The first solution to be introduced was based on bootstrapping~\cite{Sung1996} (now often named as Hard Negative Mining~\cite{girshick2014_rcnn}). This approach performs an iterative training, which selects a set of `hard' negatives, i.e. negative examples that are difficult to classify. This solution is still currently adopted in the state of the art like R-CNN~\cite{girshick2014_rcnn}, and it has been adapted recently to be used during back-propagation~\cite{Shrivastava2016, girshick15_fastrcnn}.
	
	More recently, a new object detector has been proposed (i.e. RetinaNet~\cite{Lin2017}). It is a CNN based approach in which a novel loss function, called Focal Loss, is adopted for training end-to-end and deal with class imbalance. 
	This new function is designed to down-weight easy negative examples such that their contribution to the total loss is small even if their number is large. 
	
	These solutions are effective and produce impressive results. However they are intrinsically slow because they rely on back-propagation, while bootstrapping requires to iteratively visit all negative examples in the training set. Training such systems takes hours even for medium scale datasets of few thousands of images, on servers equipped with powerful GPUs as the one used for the experimental evaluation of this work (an NVIDIA(R) Tesla P100 GPU).
	
	In this work we propose a novel approach to (i) select hard negatives, by implementing an approximated and faster bootstrapping procedure, and (ii) account for the imbalance between positive and negative regions by relying on a N\"{y}strom based kernel method (namely FALKON~\cite{falkon2018}).
	
	%%%%%%%%%%%%%%%%%%%%%%%%%%%%%%%%%%%%%%%%%%%%%%%%%%%%%%%%%%%%%%%%%%%%%%%%%%%%%%%%
	
	\section{METHODS}
	\label{sec:methods}
	
	We considered the scenario in which a human teaches the robot to detect a set of novel object instances (\textsc{TARGET-TASK} in the following). We suppose that the visual system of the robot has been previously trained on a different set of objects (\textsc{PREV-TASK} in the following), and that the convolutional weights are detained for future use.

	We propose an object detection method that can be trained on-the-fly (i.e. at run time) on the \textsc{TARGET-TASK}, by relying on some of the components previously trained on \textsc{PREV-TASK}. Referring to a typical detection pipeline, composed of (i) a region proposal stage, (ii) a feature extraction stage and (iii) a final classification stage, we propose an approach which learns (i) the region proposals and (ii) the feature extractor, on \textsc{PREV-TASK}, such that, when the robot is required to learn the \textsc{TARGET-TASK}, only the final classification stage is re-trained on-line. Our major contribution consists in an efficient approach to perform the latter step, which remarkably reduce training time while retaining performance. Specifically, we propose to use a set of FALKON classifiers, Regularized Least Squares (RLS) regressors for bounding boxes refinement, and an approximated method for selecting positive and negative samples. 
	
	\begin{figure}[htbp]
		\centering
		\vspace{3mm}
		\hspace{3mm}
		\includegraphics[width=0.47\textwidth, height=0.18\textheight]{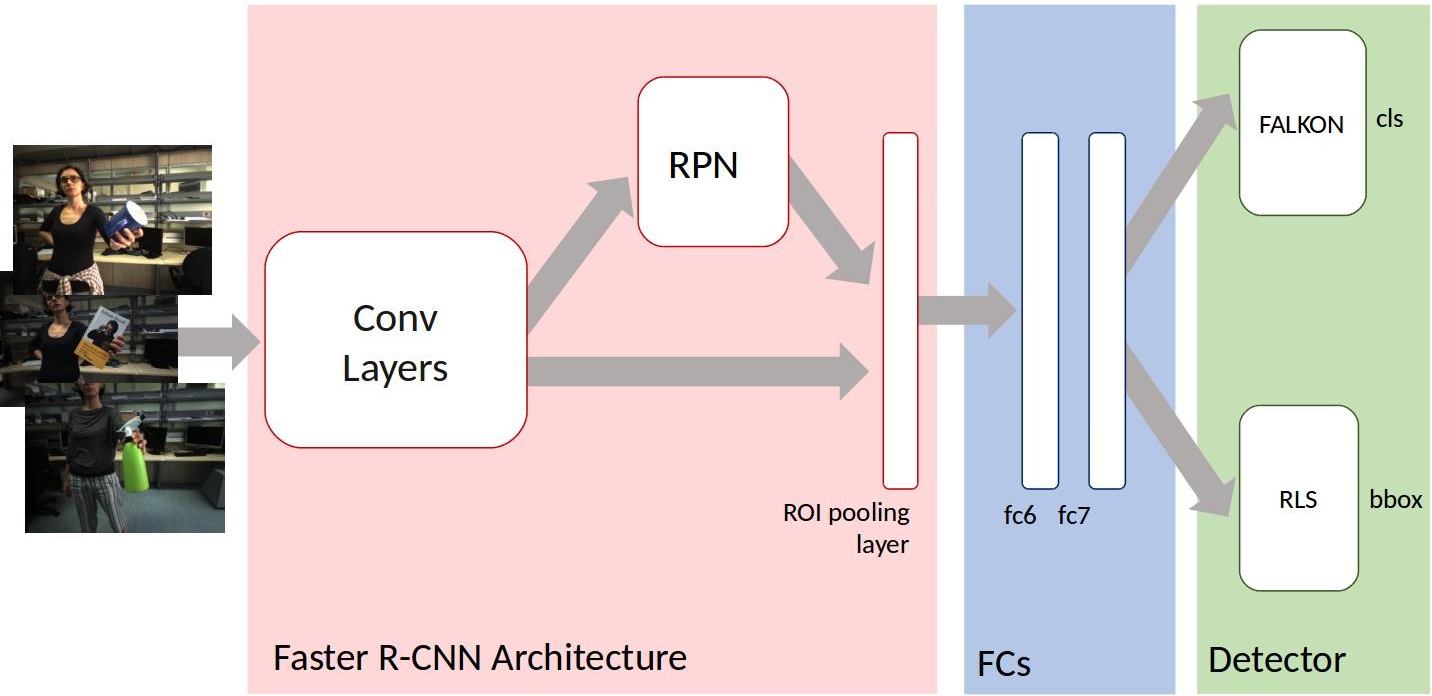}        
		\caption{Overview of the proposed on-line object detection pipeline. For a detailed description see Sec.~\ref{sec:methods}.}
		\label{fig:video}
	\end{figure}
	
	In this Section, we describe the main steps of the proposed pipeline. We first motivate the choices of the region proposal and feature extraction method (Sec.~\ref{subsec:method:pyfaster}). Then, we focus on the classifier algorithm (Sec.~\ref{subsec:method:FALKON}), and describe how it addresses the large imbalance between positive and negative examples. Finally we describe the procedure for selecting background samples (Sec. ~\ref{subsec:method:BkgSelection}).
	
	\subsection{Pipeline Overview}
	\label{subsec:method:pyfaster}
	
	We refer to Fig.~\ref{fig:video} for a pictorial representation of each stage of the pipeline. For the first stages, we adopt the architecture of Faster R-CNN~\cite{ren2015_faster}. We exploit the class-agnostic Region Proposal Network (RPN) to predict a set of candidate RoIs. Each of them is then associated to a deep feature map by means of the so-called RoI pooling layer~\cite{girshick15_fastrcnn}. This allows to encode each proposed region into a deep representation, using only one forward pass of the convolutional feature extraction layers. Specifically, we adopt the CNN model proposed in~\cite{matthew2013_zf} as convolutional backbone of the algorithm (whose integration in Faster R-CNN is publicly available\footnote{\url{https://github.com/rbgirshick/py-faster-rcnn/tree/master/models/pascal_voc/ZF}}). 
	
	Finally, in Faster R-CNN, the pooled features from the region proposals are processed by the so-called detection network, which is composed of two fully-connected layers and two final output layers for class prediction and bounding box refinement. In our pipeline, we keep the two fully connected layers, but replace the two output layers respectively with the FALKON classifiers and Regularized Least Squares (RLS) regressors for the refinement of the bounding boxes.
	
	\subsection{Training}
	
	In the considered scenario, we train Faster R-CNN on the \textsc{PREV-TASK} in order to learn the RPN and the convolutional and fully connected layers (CNN feature extractor). When learning the \textsc{TARGET-TASK}, we use these components to extract and encode region proposals into deep features, and we train on-line only the FALKON classifiers and RLS regressors.
	
	To learn the RPN and the CNN feature extractor, we adopted the 4-Steps Alternating Training method proposed in~\cite{ren2015_faster}. This method alternates the optimization of the RPN and the detection network, thus learning shared convolutional features. 
	In the first two steps, respectively the RPN and the Detector are learned from scratch, while the shared convolutional layers and the fully-connected layers are fine-tuned, starting from weights previously trained on ImageNet (the image classification task of Large-Scale Visual Recognition Challenge (ILSVRC) 2012~\cite{imagenet}). In the two latter steps, the shared convolutional layers are frozen, and the RPN and the detection network are fine-tuned on the target detection task (in our section {\sc PREV-TASK}). We refer the reader to~\cite{ren2015_faster} for a detailed explanation of the architecture and this training procedure.
	
	In the following section, we describe more in detail the algorithm and learning of the remaining part of the pipeline, which is the core of the proposed approach.
	
	\subsection{FALKON: efficient regions classification}
	\label{subsec:method:FALKON}
	
	The candidate RoIs extracted by the RPN represent potential object locations, which need to be classified as belonging to either one of the considered classes (namely, the object instances of our \textsc{TARGET-TASK}), or rejected as background.
	
	We address a multiclass classification task by learning a set of binary classifiers, one for each class, in a one-vs-all fashion. For each category, we collect the training set by selecting and labeling candidate RoIs as either positive samples (i.e. belonging to the class, indicated as $P$ in the following) or negative ones (i.e. background or other classes, indicated as $N$ in the following). This results in a large dataset. 
	
	Training standard kernel methods for classification on large datasets can be prohibitive, because it requires to solve the linear system (also known as Kernel Ridge Regression or KRR): $\left ( K_{nn}+\lambda n I \right )\alpha = \widehat{y}$,
	where $n$ is the number of training points $\{(x_1,y_1),...,(x_n,y_n)\}$, $(K_{nn})_{ij}=K(x_i,x_j)$ is the Kernel matrix and $\lambda$ is a regularization parameter. 
	
	It can be easily observed that this problem does not scale well with the number of samples $n$. Just storing $K$ requires $O(n^2)$ in memory space while computing and inverting $K_{nn}$ (i.e. learning phase) requires $O(n^2c+ n^3)$ in time (where $c$ is the kernel evaluation cost). FALKON~\cite{falkon2018} approximates the KRR problem using a N\"ystrom method~\cite{Williams2000, Smola2000} by stochastically sampling a subset of $M \ll n$ training points as Kernel centers. In addition it uses the conjugate gradient method~\cite{Saad2003} associated with a preliminary preconditioning, for an iterative and faster solution of the associated linear system. 
	
	FALKON requires $O(M^2)$ in memory space and $O(nMt+M^3)$ for the kernel computation and inversion, where $t$ is the number of iterations required. Since it has been shown that choices of $M \ll n$ preserve statistical properties~\cite{falkon2018}, accuracy is retained, while we gain a significant boost in computation performance.
	
	We refer the reader to~\cite{falkon2018} for a detailed description of the method. In our pipeline, we adopted the publicly available FALKON implementation\footnote{\url{https://github.com/LCSL/FALKON_paper}}.
	
	\subsection{Background samples selection}
	\label{subsec:method:BkgSelection}
	
	We propose two strategies in order to (i) select hard negatives and (ii) account for the imbalance between positive and negative regions. As it will be shown in the experimental Sec.~\ref{sec:experiments}, adopting these strategies in combination with FALKON proved to be fundamental in order to reduce the training time without losing accuracy.\\
	
	\noindent{{\bf Approximated Hard Negatives Mining.}} A first component of our approach is an approximation of the Hard Negatives Mining method adopted in~\cite{girshick2014_rcnn} and proposed in~\cite{Sung1996}. 
	Core idea behind the original method is to gradually grow (bootstrap) the set of negative examples by repeatedly training and testing the detector and by including in the training set only those samples which are hard to classify correctly for the detector.
	
	This idea is implemented with an iterative procedure which, for each class, visits all images in the training set and, for each image $i$: 
	\begin{enumerate}
		\item {tests the model trained at the previous iteration ($model_{i-1}$) on {\it all} negative regions in the $i^{th}$ image, to select the hard ones, maintaining a number of $N_{i}^{H}$;} 
		\item {learns a new model ($model_i$) on the train set composed by the union of the $P$ positives (which are fixed) and the hard negatives collected so far ($N_{chosen\_i-1} \bigcup N_{i}^{H}$);}
		\item {tests $model_i$ on the negatives on which it has been trained in order to prune the easy ones maintaining a number of $N_{chosen\_i}$.}
	\end{enumerate}
	The output of this procedure is a number of $N_{chosen\_final}$ (hard) negatives examples, which, jointly with the $P$ positive regions, are used to train the final version of the model. The procedure is repeated for all the binary classifiers that are trained on the multiclass problem.
	
	Such an approach is clearly time consuming, because it iterates over all images in the training set and processes {\it all} negative regions proposed by the RPN. Therefore, in our pipeline we propose to use an approximation, which consists in (i) considering a random subset of {\it all} negative regions extracted by the RPN from all training images and (ii) dividing the sampled regions into a number $n_B$ mini-batches (of size $B$). Finally we select a number of hard negatives by applying the Hard Negatives Mining method, described above, on each batch.
	
	In Sec.~\ref{sec:experiments} we compare three variants of the proposed approach:\\
	{\sc FALKON + Mini Bootstrap ($n_B \times B$):} This is the approximated procedure we outlined above, when both the size of the mini-batches $B$ and the number of bootstrapping iterations $n_B$ are parameters that can be varied to tune the procedure.\\
	{\sc FALKON + Full Bootstrap:} This is the strategy adopted by~\cite{girshick2014_rcnn}. It can be seen as performing our approximated procedure when considering all negative regions in the training set, setting each mini-batch to contain all the negatives from each of the $n$ images of the dataset (i.e., $n_B=n$ and $B$ set as the number of negatives in each image).\\
	{\sc FALKON + Random BKG:} In this case we do not perform any type of hard negatives selection, but we randomly sample a subset from all the negative regions proposed by the RPN. This can be seen as performing our approximated procedure with $n_B=0$ and $B$ set as the size of the selected subset.\\
	
	\noindent{{\bf Rebalancing N\"{y}strom centers.}} The second component of our approach is the stochastic sampling of the N\"{y}strom centers performed by FALKON, to account for the positive-negative imbalance. Specifically, we propose to condition the stochastic sampling of the $M$ centers in the algorithm, at each iteration of the approximated bootstrapping procedure described above, such that we take a number of $P'$ positives with $P'=	\min \left (P, \frac{M}{2}\right )$, while we randomly choose the remaining $(M-P')$ centers among the $N_{chosen\_i-1} \bigcup N_{i}^{H}$ negatives obtained at the $i^{th}$ iteration.
	This step is fundamental because, when $P \ll N$, randomly sampling the $M$ centers among the union of the two sets might lead to further reducing the number of positives with respect to the number of negatives and, in the worst case, discarding all positives from the sampled N\"{y}strom centers.\\
	
	The fundamental parameters of the proposed approach are (i) the number of selected N\"{y}strom centers ($M$), (ii) the number of bootstrapping iterations ($n_B$), (iii) the size of the mini-batches of negatives ($B$), and (iv) FALKON's Gaussian kernel parameters $\lambda$ and $\sigma$. We cross-validated these latter two using a one-fold cross-validation strategy, considering as validation set a subset of $20\%$ of the training set. In Sec.~\ref{sec:experiments}, we provide experimental evaluation of the other parameters which allow to tune the procedure to subsample more or less extensively the training set, depending on the desired computation time.
	
	%%%%%%%%%%%%%%%%%%%%%%%%%%%%%%%%%%%%%%%%%%%%%%%%%%%%%%%%%%%%%%%%%%%%%%%%%%%%%%%%
	\section{EXPERIMENTS}
	\label{sec:experiments}
	
	In this section we provide experimental evaluation of the proposed on-line object detection pipeline. We compare the three different training protocols described in Sec.~\ref{subsec:method:BkgSelection} using Faster R-CNN as baseline. 
	
	In Sec.~\ref{subSec:exp:icubWorld} we test the pipeline in a robotic setting using the~\textsc{iCubWorld Transformations} dataset~\cite{pasquale2016}, considering the scenario described in Sec.~\ref{sec:methods}.
	
	All experiments reported in this paper have been performed on a machine equipped with Intel(R) Xeon(R) E5-2690 v4 CPUs @2.60GHz, and a single NVIDIA(R) Tesla P100 GPU. We set FALKON to not use more than $10$GB of RAM.
	
	\subsection{Experimental Validation on PASCAL VOC 2007}
	\label{subSec:exp:pascal}
	We validate our method on a subset of PASCAL VOC 2007~\cite{pascal2010}, a standard computer vision dataset. We report performance in terms of (i) mAP (mean Average Precision), as defined for PASCAL VOC 2007, and (ii) training time.  
	\begin{table}[htp]
		\vspace{3mm}
		\centering
		\begin{adjustbox}{max width=0.99\linewidth}
			\begin{tabular}{|l|l|l|}
				\hline
				\textbf{Method}         & \textbf{mAP [\%]} & \textbf{Train Time} \\ \hline
				Faster R-CNN            & 51,9         & $\sim$25 min           \\ \hline
				FALKON + Full Bootstrap ($\sim1$K$\times$1000) & 51,5         & $\sim$8 min               \\
				FALKON + Random BKG ($0 \times 7000$)    & 47,7         & $\sim$25 sec              \\ \hline
			\end{tabular}
		\end{adjustbox}
		\caption{Performance comparison on a subset of $7$ classes of the PASCAL VOC 2007. See the text (Sec.~\ref{subSec:exp:pascal}) for details about the task and Sec.~\ref{subsec:method:BkgSelection} for details about the methods.}
		\label{table:VOC_reduced}
	\end{table}
	Specifically, we considered 7 classes among the 20 that are available (\textit{aeroplane, bicycle, bird, boat, bottle, bus, car}). As train and test set we selected, from the VOC 2007 \textit{trainval} and \textit{test} sets, all the images that depict at least one instance which belongs to one of the 7 classes. Overall we collected a training set of $\sim$1K images and a test set of $\sim$2K.\\
	
	\noindent{{\bf Approximated Hard Negatives Mining.}} As a first validation, we explored different configurations of the \textsc{FALKON + Mini Bootstrap} approach proposed in Sec.~\ref{subsec:method:BkgSelection}. We vary the number of bootstrapping iterations ($n_B$) between $0$ (i.e., no bootstrapping) and $1$K (i.e., full bootstrapping, by visiting all training images one by one), with varying size of background batches ($B$). Coherently, we observed an improvement in mAP (with an increasing training time) when progressively performing a more extensive bootstrap.
	
	We report, in Table~\ref{table:VOC_reduced}, the results provided by two ``extreme'' training conditions, comparing them with the baseline represented by fine-tuning Faster R-CNN's last layers (which we consider the ``upper bound'' for the expected mAP). 
	We consider (i) \textsc{FALKON + Random BKG}, for which we did not perform bootstrapping and set the number of randomly sampled background regions to $7000$ (higher values did not improve performance), and (ii) \textsc{FALKON + Full Bootstrap}. In this latter case, as explained in Sec.~\ref{subsec:method:BkgSelection}, we performed as many bootstrapping iterations as the number of training images (1K), processing a batch of $\sim$1000 background regions for each visited image. 
	In both experiments, in this case, we set the number of N\"{y}strom centers equal to the number of training points (this parameter is investigated in more details in the next paragraph).
	
	When fine-tuning Faster R-CNN, we froze all layers (RPN and CNN's layers up to $fc7$), except the last fully connected layers for classification and bounding box regression (namely, $cls$ and $bbox$-$reg$) which we trained from scratch. For this step we set the number of iterations to $15K$.
	
	As it can be observed from Table~\ref{table:VOC_reduced}, we could train a detection model in $\sim$25 seconds, with a $4\%$ gap in performance with respect to the mAP provided by fine-tuning Faster R-CNN's last layers (which however requires $\sim$25 minutes). Moreover, we were able to reproduce state of the art performance (up to $0.4\%$) in 8 minutes. In Sec.~\ref{subSec:exp:icubWorld}, we show that, on the robot, it is possible to find a bootstrapping configuration that remarkably reduce training time while retaining performance.\\
	
\begin{figure*}[htbp]
		\centering
		\subfloat{
			\includegraphics[width=0.29\hsize]{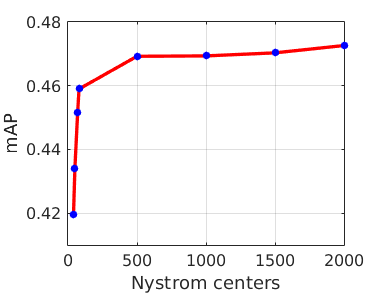}
			\label{fig:M-map_variation}
		}
		\centering
		\subfloat{
			\centering
			\includegraphics[width=0.29\hsize]{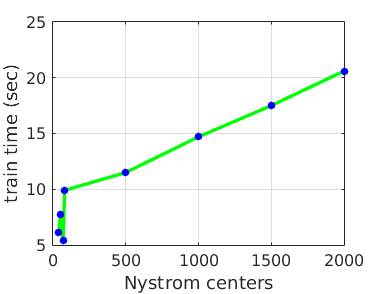}
			\label{fig:M-train-time_variation}
		}
		\centering
		\subfloat{
			\centering
			\includegraphics[width=0.29\hsize]{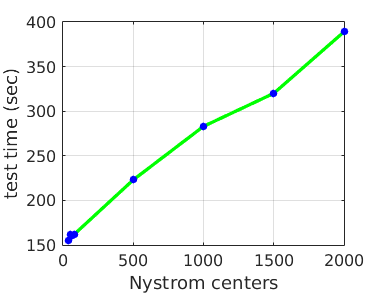}
			\label{fig:M-inference-time_variation}
		}
		\caption{The three plots refer to the experiment discussed in Sec.~\ref{subSec:exp:pascal} and report the mAP (Left), training time (Center) and testing time (Right) for increasing values of the number of N\"{y}strom's centers ($M$) when using FALKON.}
		\label{fig:M-variation}
	\end{figure*}

	\noindent{{\bf Rebalancing N\"{y}strom centers.}} As a second validation, in Fig.~\ref{fig:M-variation} we report performance on the same task when varying the number of N\"{y}strom centers ($M$ parameter in Sec.~\ref{subsec:method:FALKON}). For this experiment, we considered the \textsc{FALKON + Random BKG} configuration of Table~\ref{table:VOC_reduced}, and progressively decreased $M$, by applying, for each value, the rebalancing approach described in Sec.~\ref{subsec:method:BkgSelection}.
	For each value of $M$, we report the mAP (Left), the training time (Center) and the testing time (Right).
	It can be observed that mAP performance degrades only when using very few centers, and that a value of $M=\sim500$ is sufficient to achieve optimal performance, fast training ($\sim$12 seconds) and testing time (around $\sim$10 FPS, obtained by dividing the number of tested images, $\sim$2K, by the reported testing time for $M=500$, i.e., $\sim$225 seconds).
	\subsection{Experiments on the \textsc{iCubWorld Transformation} dataset}
	\label{subSec:exp:icubWorld}
	To evaluate the effectiveness of the proposed pipeline in a robotic scenario, we consider the \textsc{iCubWorld Transformations} dataset (\textsc{iCWT} in the following).\\
	
	\noindent{{ \bf iCubWorld Transformations.}} This dataset has been automatically acquired in a natural teacher-learner setting by using the iCub robot~\cite{icub}. The teacher shows an object in front of the robot, which moves the eyes to fixate and track it. Using its stereo system, the robot segments the object by selecting those points which are closest to the cameras\footnote{This approach assumes that the  object of interest is the closest entity to the robot in the scene, which in practice holds in our scenario}. The bounding box that contains the segmented object, is then stored jointly with the label of the object which is provided verbally by the teacher. We refer to~\cite{pasquale2016_frontiers} for a complete description of the acquisition setup and to~\cite{pasquale2016} for details about the dataset, which is publicly available\footnote{\url{https://robotology.github.io/iCubWorld/}}.\\
	
\begin{figure*}[htbp]
		\centering
		\vspace{3mm}
		\includegraphics[width=0.99\textwidth]{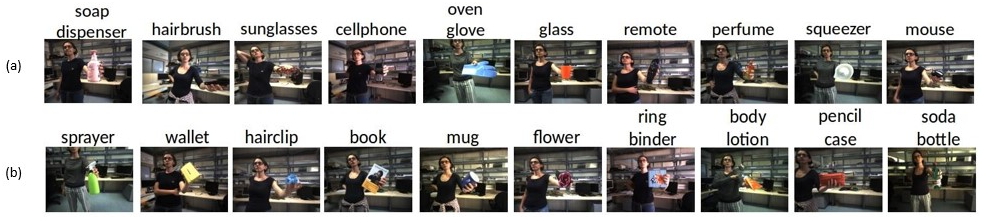}        
		\caption{Example images representing the $10$ object instances involved  in the first object identification task (\textsc{PREV-TASK}), i.e., the task on which we learn the RPN and the CNN feature extractor (a) and in the second object identification task (\textsc{TARGET-TASK}), i.e., the actual task that is learned on-line (b). }
		\label{fig:iCWTdataset}
	\end{figure*}

	\noindent{{\bf Experimental Setup.}} For our experiments we considered the scenario described in Sec.~\ref{sec:methods} and we defined the two tasks (\textsc{PREV-TASK} and \textsc{TARGET-TASK}) as two object identification tasks among $10$ object instances. We randomly chose one object instance from each of the $20$ different categories in the dataset, and split them into two sets of $10$ objects each. We show an example image for each object considered in both tasks in Fig.~\ref{fig:iCWTdataset}.
	
	For each task, we considered, as training set for each object, the union of the $4$ image sequences available in {\sc iCWT}, corresponding to the {\sc 2D ROT},  {\sc 3D ROT},  {\sc BKG} and  {\sc SCALE} viewpoint transformations, taking into account both acquisition days. Overall this leads to a set of  $\sim$10K images. As a test set for each object we used the remaining {\sc MIX} sequence, in both acquisition days. We refer to~\cite{pasquale2016} for details about the dataset's sequences.
	
	We trained Faster R-CNN end-to-end on \textsc{PREV-TASK}, setting the number of iterations to $40$K when learning the RPN and to $20$K when learning the CNN detector. 

	We finally trained the {\sc FALKON} classifiers and evaluated performance on the test set of \textsc{TARGET-TASK}.
	
	As for PASCAL VOC, we considered Faster R-CNN as a baseline. To this end, we used the model obtained by training end-to-end on \textsc{PREV-TASK}, and froze all layers (RPN and CNN's layers up to $fc7$), while learning from scratch only the last fully-connected layers for classification and bounding box regression (namely, $cls$ and $bbox$-$reg$). For this step we set the number of iterations to $20$K.\\

	\noindent{{\bf Results.}} In Table~\ref{tab:iCWT-results} we compare the performance of Faster R-CNN baseline with the methods \textsc{FALKON + Random BKG} and \textsc{FALKON + Mini Bootstrap} (Sec.~\ref{subsec:method:BkgSelection}).
	
	Based on the empirical observations from Sec.~\ref{subSec:exp:pascal}, in these experiments we use a relatively large value of N\"{y}strom centers (around half the size of the training set), because this does not have a negative impact on the training time as demonstrated by the experiments in Table~\ref{table:VOC_reduced}. 
	
	As in PASCAL VOC, for \textsc{FALKON + Random BKG} we set the number of randomly sampled background regions to $6000$, because we observed that further increasing this parameter did not improve performance.
	
	Regarding \textsc{FALKON + Mini Bootstrap}, we observed that a few bootstrapping iterations already give  performance that are comparable to the Faster R-CNN. As an example, in Table~\ref{tab:iCWT-results} we report two different configurations of  \textsc{FALKON + Mini Bootstrap}: (i) $n_B=4$ and $B=2500$ (\textsc{FALKON + Mini Bootstrap ($4\times2500$)}) and (ii) $n_B=10$ and $B=1500$ (\textsc{FALKON + Mini Bootstrap ($10\times1500$)}).
	
	It can be noticed that \textsc{FALKON + Random BKG} has the fastest training time but with lowest mAP. On the contrary, the proposed bootstrap method, \textsc{FALKON + Mini Bootstrap ($4\times2500$)} provides mAP performance which are comparable with those obtained with Faster R-CNN, with just 40 seconds of training (in contrast fine-tuning Faster R-CNN takes 40 minutes). Moreover, \textsc{FALKON + Mini Bootstrap ($10\times1500$)} outperforms fine-tuning with a train time of $\sim 50$ seconds.
	
	\begin{table*}[htbp]
		\centering
		\begin{adjustbox}{max width=0.99\linewidth}
			\begin{tabular}{|l|l|l|llllllllll|}
				\hline
				\textbf{Method}                   &\begin{tabular}[c]{@{}l@{}}\textbf{Train}\\ \textbf{Time}\end{tabular}       & \textbf{mAP [\%]} & \begin{tabular}[c]{@{}l@{}}soda\\ bottle\end{tabular} & mug  & \begin{tabular}[c]{@{}l@{}}pencil\\ case\end{tabular} & \begin{tabular}[c]{@{}l@{}}ring\\ binder\end{tabular} & wallet & flower & book & \begin{tabular}[c]{@{}l@{}}body\\ lotion\end{tabular} & \begin{tabular}[c]{@{}l@{}}hair\\ clip\end{tabular} & sprayer \\ \hline
				Faster R-CNN Fine-tuning           & $\sim$40 min           & 49,7             & 63,2       & 68,4 & 23,3       & 29,6       & 49,9   & 66,1   & 35,3 & 56,2       & 60,2     & 45,8    \\ \hline
				FALKON + Random BKG  (0$\times$6000)             & \textbf{$\sim$25 sec} & 40,5             & 57,7       & 67,9 & 17,5       & 23,1       & 23,8   & 59,5   & 26,8 & 39,6       & 48,5     & 40,5    \\
				FALKON + Mini Bootstrap (4$\times$2500)  & $\sim$40 sec          & 48,1             & 63,1       & 67,2 & 18,4       & 25,7       & 47,4   & 70,3   & 36,1 & 52,3       & 58,8     & 41,5    \\
				FALKON + Mini Bootstrap (10$\times$1500) & $\sim$50 sec          & \textbf{51,3}    & 64,7       & 71,2 & 27,2       & 31,7       & 56,9   & 69,4   & 39,6 & 54,0       & 60,7     & 37,1    \\ \hline
			\end{tabular}
		\end{adjustbox}
		\caption{Performance comparison on a $10$-object identification task on {\sc iCWT} ({\sc TARGET-TASK} in Sec.~\ref{subSec:exp:icubWorld}). See the text for details about the task and Sec.~\ref{subsec:method:BkgSelection} for details about the methods.}
		\label{tab:iCWT-results}
	\end{table*}
	%%%%%%%%%%%%%%%%%%%%%%%%%%%%%%%%%%%%%%%%%%%%%%%%%%%%%%%%%%%%%%%%%%%%%%%%%%%%%%%%
	\section{CONCLUSIONS}
	
	In this paper we propose a system for object detection that combines Faster R-CNN~\cite{ren2015_faster}, with FALKON~\cite{falkon2018}, a kernel-based method specifically designed for large-scale datasets. In addition, we include a novel approximated technique for pruning the training set by selecting hard negative examples. 
	Our approach can be trained much faster than Faster R-CNN ($\approx 60 \times$) while preserving comparable detection performance. 
	
	Fast learning methods are fundamental for robots to quickly adapt to their environment. Deep-learning methods for object detection achieve remarkable performance, but are slot to train. This hinder their adoption in those scenarios that require robots to learn on-line. In this sense the work described in this paper is an important step toward the implementation of more adaptive robotic systems. 
	
	%%%%%%%%%%%%%%%%%%%%%%%%%%%%%%%%%%%%%%%%%%%%%%%%%%%%%%%%%%%%%%%%%%%%%%%%%%%%%%%%
	
	\section*{ACKNOWLEDGMENTS}
	
	The authors would like to thank NVIDIA Corporation for the donation of a Tesla K40c GPU used for this research. This work is funded by the Air Force project FA9550-17-1- 0390 (European Office of Aerospace Research and Development) and by the MSCA- RISE, 734564.

	\normalsize
	\bibliographystyle{plain}
	\bibliography{bibliography}

\end{document}